\documentclass[10pt,journal]{IEEEtran}

\usepackage{amsmath,amssymb,amsfonts}
\usepackage{algorithmic}
\usepackage{graphicx}
\graphicspath{{figures/}}
\usepackage{textcomp}
\usepackage{xcolor}

\usepackage[utf8]{inputenc}
\usepackage[hyphens]{url}
\usepackage{hyperref}
\usepackage{cleveref}
\usepackage{booktabs}
\usepackage{float}
\usepackage{xspace}
\usepackage{subfigure}
\usepackage{multirow}
\usepackage{csquotes}
\MakeOuterQuote{"}

\newcommand{\PAR}[1]{
	\noindent\textbf{#1}}

\ifCLASSOPTIONcompsoc
  \usepackage[nocompress]{cite}
\else
  \usepackage{cite}
\fi

\ifCLASSINFOpdf
\else
\fi

\begin{document}
\title{Deep Traffic Sign Detection and Recognition Without Target Domain Real Images}

\author{
	\IEEEauthorblockN{Lucas Tabelini,
		Rodrigo Berriel,
		Thiago M. Paix\~ao,
		Alberto F. De Souza,\\
		Claudine Badue,
		Nicu Sebe and
		Thiago Oliveira-Santos}
\IEEEcompsocitemizethanks{
	\IEEEcompsocthanksitem L. Tabelini, R. Berriel, C. Badue, A. F. De Souza, and T. Oliveira-Santos are with Universidade Federal do Espirito Santo (UFES), Brazil.\protect\\
	\IEEEcompsocthanksitem T. M. Paix\~ao is with Instituto Federal do Espirito Santo (IFES), Serra, Brazil, and Universidade Federal do Espirito Santo (UFES), Brazil.\protect\\
	\IEEEcompsocthanksitem Nicu Sebe is with University of Trento, Trento, Italy.
}
}

\maketitle
\begin{abstract}
	Deep learning has been successfully applied to several problems related to autonomous driving, often relying on large databases of real target-domain images for proper training. The acquisition of such real-world data is not always possible in the self-driving context, and sometimes their annotation is not feasible. Moreover, in many tasks, there is an intrinsic data imbalance that most learning-based methods struggle to cope with. Particularly, traffic sign detection is a challenging problem in which these three issues are seen altogether. To address these challenges, we propose a novel database generation method that requires only (i) arbitrary natural images, i.e., requires no real image from the target-domain, and (ii) templates of the traffic signs. The method does not aim at overcoming the training with real data, but to be a compatible alternative when the real data is not available. The effortlessly generated database is shown to be effective for the training of a deep detector on traffic signs from multiple countries. On large data sets, training with a fully synthetic data set almost matches the performance of training with a real one. When compared to training with a smaller data set of real images, training with synthetic images increased the accuracy by 12.25\%. The proposed method also improves the performance of the detector when target-domain data are available.
\end{abstract}


\IEEEdisplaynontitleabstractindextext

\IEEEpeerreviewmaketitle

\section{Introduction}\label{sec:introduction}

\IEEEPARstart{D}{eep} neural networks (DNNs) have been widely used to tackle a variety of computer vision tasks,  particularly on several problems related to autonomous driving~\cite{oursurvey2019arxiv}. Many of these applications rely on large networks which usually require large amounts of data to be properly trained. This requirement, however, is not always easy to be fulfilled. Acquiring problem-specific real-world databases, especially in robotics, is often a hard task, particularly when considering the additional annotation process. In this context, it is desirable to produce high-performance models without the need of annotated real-world images. %

The success of deep learning on autonomous driving and on advanced driver assistance systems (ADAS) has been demonstrated on several applications: scene semantic segmentation~\cite{deeplab2018eccv}, traffic light detection~\cite{possatti2019traffic}, crosswalk classification~\cite{berriel2017cag}, traffic sign detection~\cite{zhu2016cvpr}, pedestrian analysis~\cite{sarcinelli2019pedestrian}, car heading direction estimation~\cite{DBLP:conf/ijcnn/BerrielTCGBSO18} and many other applications. This work focuses on the traffic sign detection problem, whose goal is to locate signs of interest along the road (with the help of a camera mounted on a vehicle) and classify their specific type (e.g., whether it is a 60 or a 80 km/h sign). This is an important task to be performed by autonomous driving systems and ADAS because traffic signs set rules which (i) the drivers are expected to abide by and (ii) road users should rely on while making decisions.

The traffic sign detection problem has been investigated by the research community for a while. Researchers have been proposing all types of solutions such as the ones using hand-crafted features in model-based solutions~\cite{barnes2008tits},  leveraging simple features in learning-based approaches~\cite{bascon2007tits}, and, the more recent and state-of-the-art, using deep learning based methods~\cite{zhu2016cvpr, garcia2018neurocomputing} that is the focus of this work.

Apart from the major advances on the topic, there are still many issues requiring further investigation, specially when considering deep learning approaches for detection. In general, the training of deep detectors require (i) expensive annotation, (ii) real images from the target domain,
and (iii) balanced data sets.
The annotation process is expensive because each traffic sign has to be marked with a bounding box, which is more difficult than just assigning a class for an object in a classification problem.
Moreover, deep detectors are still known for being data hungry, i.e., they require many real images to perform well. Therefore, the acquisition of such images with traffic signs can be troublesome, %
because it requires finding many traffic sign samples along the roads. Since traffic legislation changes from country to country, the traffic signs are not standardized across the world and a new data set has to be created for every country.
Finally, the image acquisition process should yield a balanced number of exemplars of each class. This would require collecting many more images to have a minimum balance across the classes because some traffic signs are rarer than others under common driving circumstances, causing a long-tail effect~\cite{zhu2016cvpr}.

In this context, we hypothesize that the training of a deep traffic sign detector without annotated real images from the domain of interest (i.e., target domain real images) can yield a performance similar to those models trained on manually annotated images from the domain of interest. The method does not aim at overcoming the training with real data, but to be a compatible alternative when the real data is not available. Therefore, this work proposes a novel effortless method for generation of synthetic databases that requires only (i) arbitrary natural images (i.e., images out of the domain or context of interest) as background and (ii) templates of the traffic signs (i.e., synthetic representative images of the different traffic signs). The synthetic database is used to train a country-optimized deep traffic sign detector. We argue that, in the context of traffic sign detection, the proposed database generation process handles the three previously mentioned issues altogether facilitating the training of country-specific detectors. In a preliminary work~\cite{tabelini2019ijcnn}, the hypothesis was verified, but only for a simplified case, as detailed next.

The traffic sign detection task can be divided in two steps: localization and recognition. Although the first step is crucial, it becomes more useful in practice when is followed by the second step. It is not enough to know that there are traffic signs in the scene, since different traffic signs convey different information. It is in the recognition step that traffic signs are distinguished, for example, to verify the speed limit on the road. In the preliminary version of this paper \cite{tabelini2019ijcnn}, the recognition step was not addressed since the focus was precisely on locating traffic signs in the scene. In that occasion, the proposed method was shown to be effective on the localization task for German traffic signs. This work extends and consolidates the preliminary investigation by including: the recognition task, an extensive experimentation, and a more in-depth discussion of the results. Although the recognition step implies significant additional complexity, experimental results show that the proposed method is successful, which makes it useful in practice. In the experimentation, two large data sets were added, both more than ten times larger than the one used in the preliminary work. To assess the impact of each step in the proposed method, an ablation study was performed. In addition, the previous work raised a question: how many real images are equivalent to using synthetic ones? This work shows that the amount of real data required to match the performance of the system trained with synthetic data only is huge for two data sets (in the order of tens of thousands). Finally, two additional use cases were investigated for the synthetic data, data augmentation and pretrain. The data augmentation investigation shows that the synthetic data together with a few real data samples can remarkably boost the performance of a detector to match those trained with a large amount of real images. The pretrain investigation shows that the proposed synthetic data generation process is useful even when a large amount of real data from the target domain is available. Results show that pretraining the detector with the synthetic data generated with our method improves the learning of the detection model and substantially increases the final accuracy.

\section{Related works}
\label{sec:related_works}
\begin{figure*}[t]
	\centering
	\includegraphics[width=\linewidth]{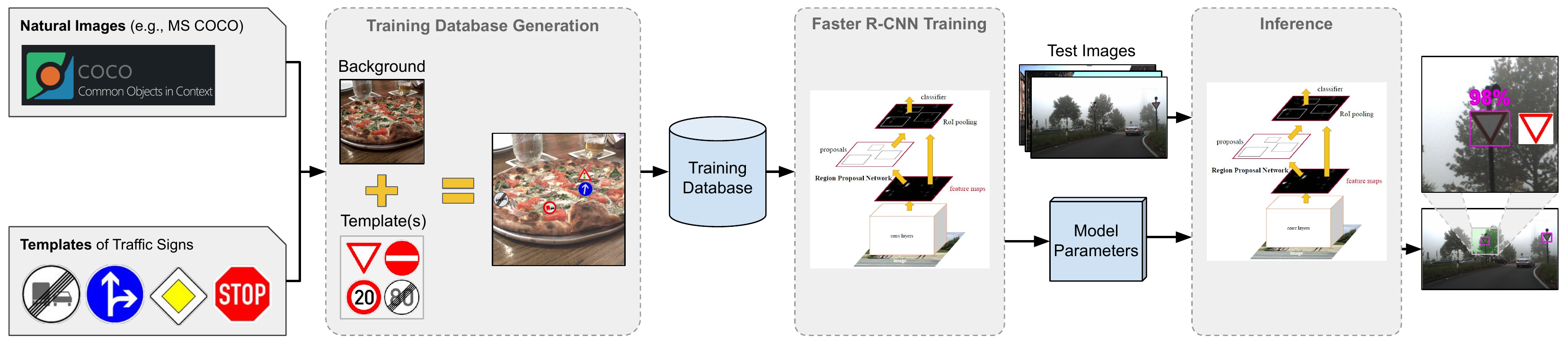}
	\caption{Overview of the proposal method. From left to right, the method receives as input natural images (e.g., from publicly available large-scale databases) and templates of traffic signs and generates a synthetic training database. The synthetic database is used to train a deep detector (e.g., Faster R-CNN). Finally, the model is ready to detect and recognize traffic signs.}
	\label{fig:overview}
\end{figure*}

This section presents a brief review of general traffic sign detection methods and of methods for generating synthetic data for deep training.

\subsection{Traffic Sign Detection}
In the past, most methods proposed in the literature to tackle the problem of traffic sign detection used classic computer vision approaches. Since traffic signs may be easily distinguished by humans due to the contrasting colors, the first works took advantage of those features~\cite{ruta2008detection}. In this context, techniques such as color thresholding are used to segment the traffic signs, followed by a post-processing step~\cite{varun2007road}. Although those methods are generally faster than others, they are not robust to factors such as weather conditions, occlusion, or time of the day \cite{saadna2017overview}. Other approaches take advantage of the well-defined shapes of traffic signs. To detect them by the shape, many methods may be used, but the most common ones employ the Hough transform~\cite{zaklouta2012real}. Although these two approaches (based on color and shape) work well separately, the results can be improved if they are combined \cite{loy2004fast}. When combined, either method can be used first, while the second is usually applied to filter the results. With the traffic signs detected, the next step is the recognition. For this step, the most common methods are neural networks \cite{carrasco2012recognition,kim2018efficient}, genetic algorithms~\cite{de2001traffic}, AdaBoost classifiers \cite{lin2012road} and SVMs \cite{fang2003road}. More recently, with data becoming easier to acquire, deep learning has also shown success on the traffic sign recognition task, particularly with the use of convolutional neural networks (CNNs) \cite{ciresan2011committee,wu_traffic_2013}. In 2010, a benchmark for traffic sign recognition was proposed~\cite{Stallkamp2012}. The best results were achieved using CNNs~\cite{ciresan_multi-column_2012}. It is important to note that in many works that tackle the traffic sign detection problem with deep learning the classification is done only to distinguish between super-categories and not between specific classes (e.g., a 70 or 80 km/h traffic sign has the same class label)~\cite{wu_traffic_2013,garcia2018neurocomputing}. The main issue in most works with deep detectors is the need for large amounts of data. In this work, we propose a method to train a deep detector with no real data and focus on distinguishing between every specific class (fine-grained class recognition). For a more in-depth review of traffic sign detection the reader can refer to \cite{wali_vision-based_2019}.

\subsection{Synthetic Data Generation for Deep Training}
Methods to generate synthetic data for deep training have been extensively studied in the past. Those methods can be divided in two groups: non-learning (i.e., there is no learning during the synthetic data generation) and learning based (i.e., there is learning during the synthetic data generation).
For the detection task, most non-learning based methods cut and paste objects from an image to another. For instance, Dwibedi et al.~\cite{dwibedi_cut_2017} cut objects with a segmentation mask and then blend them on other images from the target domain. Wang et al.~\cite{wang_data_2019} use a similar approach, where different objects from the same category switch positions. 3D models have also been used to generate data for training, overlaying 2D renderings of those models on real images \cite{peng_learning_2015}. In the task of traffic sign classification, some works used image processing to generate training samples from traffic sign templates~\cite{moiseev_evaluation_2013,stergiou_traffic_2018}. In the detection task, Møgelmose et al.~\cite{mogelmose2012learning} tried using synthetic data to train a Viola-Jones traffic sign detector, but the results were not satisfactory.
Learning based approaches are motivated by the premise that training the model with more realistic synthetic data will lead to a better performance on real-world data. The learning process can be used to, for example, generate data with objects in a more natural position, as Dvornik et al.~\cite{ferrari_modeling_2018} have shown that cutting and pasting objects at random positions may not be ideal. In particular, Georgakis et al.~\cite{georgakis_synthesizing_2017} place cropped objects of interest from public data sets on locations that are most likely to be a surface, being such positions estimated via semantic segmentation. The scale of the objects to be placed is determined according to the depth value associated to the surface position. Gupta et al.~\cite{gupta_synthetic_2016} use a similar approach for text localization, predicting a depth map of each background image. With the depth map, regions are filtered to gather suitable regions for text placement. Then, the text is superimposed on those regions, also using the depth map to determine the text's perspective. In the context of classification using one-shot learning, Grigorescu~\cite{grigorescu_generative_2018} proposes to generate data with predefined functions that make templates more realistic. To set those functions' parameters, a network is trained using templates (called one shot objects, in his work) and real traffic sign samples. Kim et al.~\cite{kim_variational_2019} propose an approach with a variational prototyping-encoder. In the training process, real training images are encoded to a latent space and then decoded to a prototype (template). In the testing phase, the encoder is used as a feature extractor and a nearest neighbor classifier is used on the features extracted from the test image and the templates.

In all the aforementioned works, the proposed method is either evaluated on the classification task only or requires annotated real-world data from the problem domain. In this work, we tackle the detection task with fine-grained class recognition using no problem domain real-world data.

\section{Proposed Method}
The proposed method (illustrated in \Cref{fig:overview}) comprises mainly the generation of a synthetic training data set that requires no real image from the domain of interest. After that, this synthetic data set is used to train a deep traffic sign detector. Finally, the trained deep detector model can be used to infer the position and the class of traffic signs on real images.
\subsection{Training Database Generation}
\label{sec:samples_generation}

The generation of the training database is three-fold. First, templates of the traffic signs of interest are acquired. Then, background images that do not belong to the domain of interest are collected (e.g., arbitrary natural images). Lastly, the training samples, comprising images with annotated traffic signs, are generated.

\vspace{.5em}
\PAR{Template acquisition.} The first step towards the generation of the training samples is the acquisition of a template for each traffic sign of interest. The traffic signs of interest are those which the system is expected to operate with. Frequently, traffic signs are part of country-wise specific legislations defined by governmental agencies. This usual country-wise standardization helps the acquisition of templates (which is the goal of this step), because their very definitions (i.e., the templates) are part of pieces of legislation commonly available on-line on the websites of these agencies. In fact, templates are graphic representations of these definitions. In addition, some of the publicly available data sets for traffic sign detection (e.g.,~\cite{Houben-IJCNN-2013}) also distribute the templates of the classes annotated in their samples. All of this makes it easy and convenient to acquire templates virtually for any given set of standardized traffic signs worldwide. In case the templates are not available online, it is always possible to draw it manually. These templates are acquired and stored to be used later.

\PAR{Background acquisition.}
In addition to the templates, the system requires images to use as background of the training samples. Although the natural choice would be images that belong to the domain of interest, e.g., images of roads, highways, streets, etc., we argue that this is not required. Moreover, we believe that choosing background images from the domain of interest may introduce unwanted noise in the training data if not carefully annotated. Images from the domain of interest eventually will present the object of interest, which, in turn, will be treated as background as well. By not constraining the background acquisition to images of the domain of interest, many of the freely available large-scale data sets (e.g., ImageNet~\cite{krizhevsky2012imagenet}, Microsoft COCO~\cite{lin2014microsoft}, etc.) can be exploited. For this work, the Microsoft COCO data set was chosen to be used as background, except for the images containing classes that are closely related to the domain of interest in order to avoid the introduction of noise in the training data set. Details of the background acquisition are presented in~\Cref{sec:experiments}. After choosing the background images, the training samples can be generated.

\vspace{.5em}
\PAR{Training samples generation.}
The last step of the training database generation is blending the background images and the templates of the traffic signs. This blending process aims to reduce the appearance difference between the background and the templates. If successful, the samples generation process is advantageous because it may tackle all the three aforementioned issues: (i) the training samples are automatically annotated, since the position of the traffic signs on the image is defined by the method; (ii) a large-scale database can be generated without much cost, given that there are a lot of different possible combinations between the random natural images (i.e., the backgrounds) and the objects of interest (i.e., the transformed traffic sign templates); and (iii) the training data set will not suffer from imbalance, since the method can sample the classes uniformly. The details of the blending process are described next.

\begin{figure}[H]
	\centering
	\includegraphics[width=0.8\linewidth]{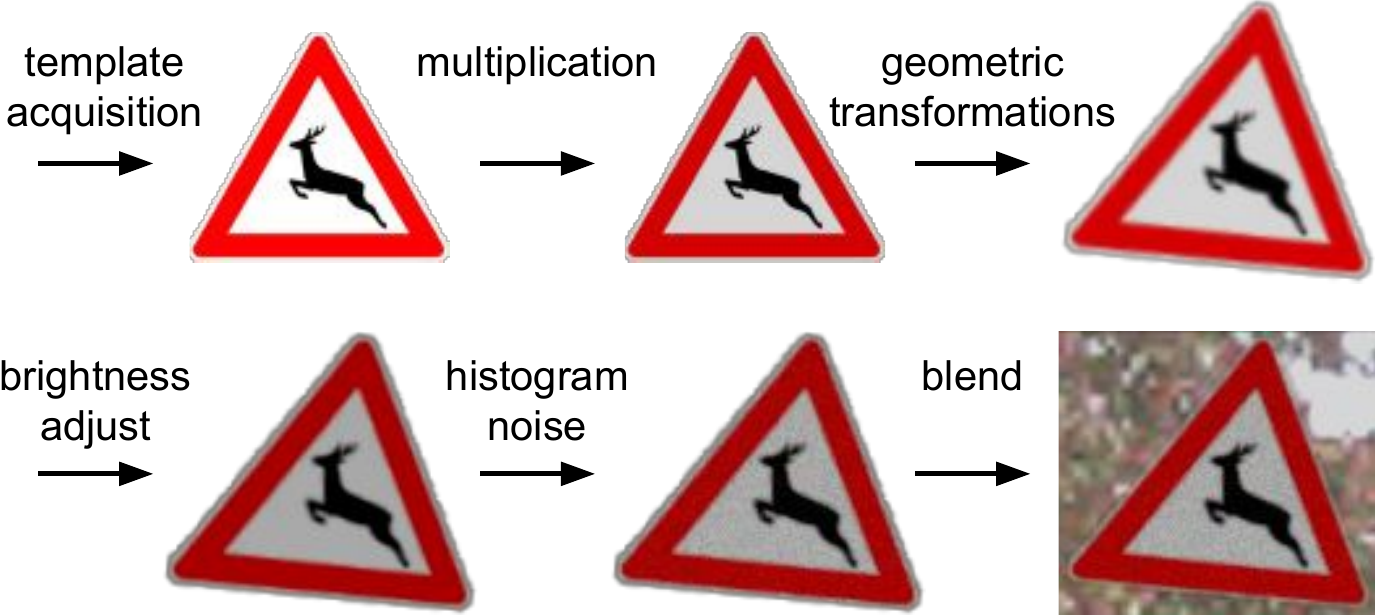}
	\caption{Steps of the template blending process.}
	\label{fig:step_by_step}
\end{figure}

\begin{figure}[t]
	\centering
	\includegraphics[width=\linewidth]{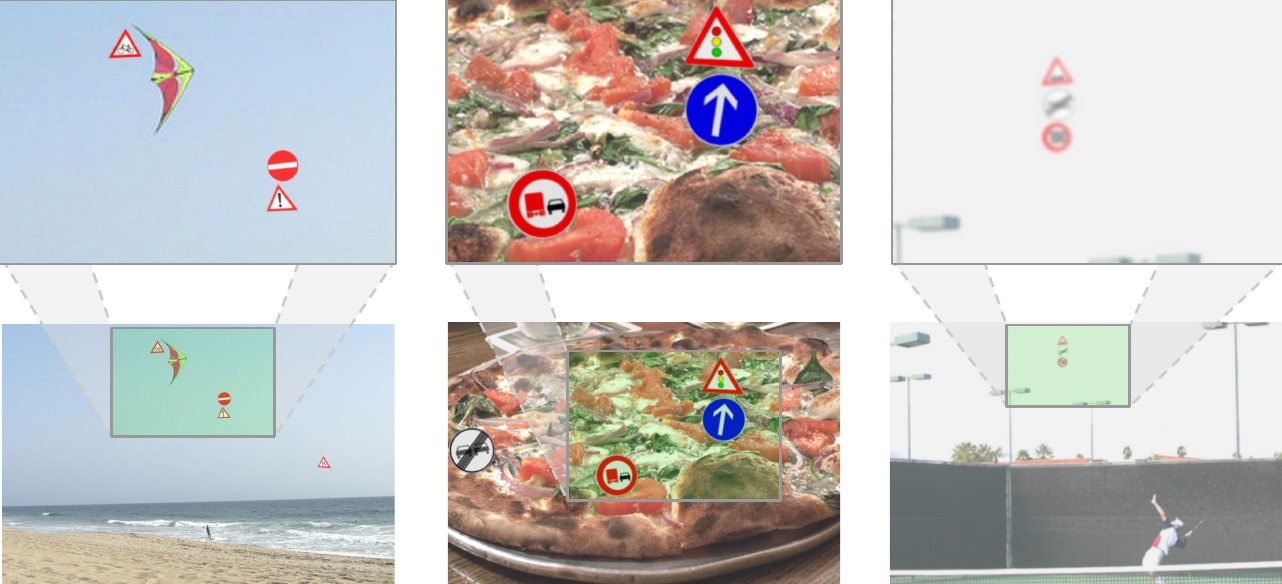}
	\caption{Some training samples can be seen in the bottom row, and zoomed-in figures highlighting the regions with traffic signs can be seen in the top row. These samples were generated using the process described in~\Cref{sec:samples_generation}.}
	\label{fig:blend_samples}
\end{figure}

Let $\mathcal{B} = \{B_{i}\}_{i=1}^{S}$ 
be the background set with $S$ random natural images, in total; 
$\mathcal{C} = \{C_{i}\}_{i=1}^{M}$ 
be the set of classes of interest; $\mathcal{T} = \{T_i \mid i \in \mathcal{C} \}$ be the set of templates of traffic signs; and $a$, $b$, $c$, $d$, $e$, and $f$ be input parameters. First, for a training data set with $N$ samples, the training set is defined as $\mathcal{X} = \{X_1, X_2, \cdots, X_N\} \stackrel{iid}{\sim} \mathcal{B}$. The first step is to randomly change the brightness and contrast of the image by randomly adding and multiplying each training sample, i.e., $\alpha_{i} X_i + \beta$, where $\alpha_{i} \sim \text{\textit{U}}(a, b)$, $\beta \sim \text{\textit{U}}(c, d)$, and $X_i$ is the background image sampled for the $i$-th training sample. The next step is to add a random amount of $|K^{i}|$ templates into the $i$-th sample $X_i$, where $|K^{i}| \in \{1, 2, \dots, M\}$ and $K^{i}_{j} \sim \mathcal{T}$. The templates are placed in random configurations (e.g., a $2\times4$ grid, $1\times3$ row, etc) from a predefined set. This step attempts to mimic a common behavior in real world (considering the country of interest), where sometimes multiple signs are seen together. The process of adding a template into a background image is as follows: (i) multiply the template $K^{i}_{j}$ by the same $\alpha_{i}$ used on the background image $X_i$; (ii) apply geometric transformations (3D rotations and scaling) (iii) adjust the brightness by adding to the template the average of the region on which the template is being added, minus a constant;
(iv) add noise, i.e., jitter; (v) place the template into a random position (unless it is tied to another template) with no intersection with the others; and (vi) fade the borders of the template to create a smooth transition from the template to the background image. Lastly, a Gaussian blur $\sigma \sim U(0, max(e, f \times \textrm{scale}))$ is applied to the resulting image, generating the final training sample. Some training samples can be seen in \Cref{fig:blend_samples}, while a step-by-step overview of the blending process can be seen in \Cref{fig:step_by_step}.

Finally, it is important to generate the templates according to the range of operation of the application. Therefore, it is important to set up the minimum and maximum size of the templates to be detected and sample the random scales accordingly. This procedure can be seen as a calibration step on which, in a real-world application, one could determine the minimum and maximum size of a traffic sign by looking at few images of a particular target camera. Models, code and the parameters used will be made available\footnote{The link will be available upon acceptance.}.

\subsection{Model Training and Inference}
\label{sec:training}
After generating the training database, a deep detector can be trained. In this work, the state-of-the-art for traffic sign detection~\cite{garcia2018neurocomputing}, Faster R-CNN~\cite{fasterrcnn} framework, was chosen. In every experiment conducted, the training model was initialized from a model pre-trained on ImageNet~\cite{krizhevsky2012imagenet}. Roughly, the Faster R-CNN is a 2-step detection framework comprising (i) the Region Proposal Network (RPN), which predicts regions that are likely to contain an object, then (ii) two fully-connected networks, one to refine the predicted regions, and the other to predict the class of each object. After being trained, the Faster R-CNN can process RGB input images of traffic scenes predicting bounding boxes, classes, and confidence scores of the predicted traffic signs.

\section{Experimental Methodology}
\label{sec:experiments}

This section introduces the data sets used to train and evaluate the detection models, the metrics for performance
quantification, and the experiments conducted to validate our proposal. The experimental platform is described at the
end of the section.
\subsection{Data sets}
A public image data set was used as source for background images, while three data sets of traffic scenes were used to evaluate the proposed method. Each data set is described in the following paragraphs.

\PAR{Backgrounds source.}
Microsoft COCO (MS-COCO)~\cite{lin2014microsoft} is a large-scale data set (more than 200k labeled images
divided into training and test sets) designed for the tasks of object detection, segmentation, and visual captioning.
For this work, the images of the 2017 version of MS-COCO are used as background for the traffic signs
templates, as described in \Cref{sec:samples_generation}. More specifically, the sign templates are overlaid onto the images of the 
MS-COCO training partition in order to train the traffic sign detector. For our purposes,
traffic-related scenes should be disregarded, which is done by filtering out those images originally labeled as
``traffic light'', ``bicycle'', ``car'', ``motorcycle'', ``bus'', ``truck'', ``fire hydrant'', ``stop sign'',
and ``parking meter''. Images with height less than 600 pixels or width less than 400 pixels are also removed, totaling at the
end 58078 images. The remaining images are further uniformly scaled so that the shortest dimension has 1500 pixels.
Finally, the central 1500 $\times$ 1500 pixels area is cropped from the scaled image.

\PAR{Evaluation datasets.}
The three datasets used to evaluate the proposed method were Belgian Traffic Signs Dataset (BTSD) \cite{timofte_multi-view_2014}, Tsinghua-Tencent 100K benchmark (TT100K)~\cite{zhu2016cvpr} and German Traffic Sign Detection Benchmark (GTSD)~\cite{Houben-IJCNN-2013}. On BTSD, although the data set has annotations for more than a hundred classes, it also provides a reduced list of 62, which was used in this work. On CTSD, originally, there are 151 traffic sign classes, however, only a subset of 42 classes is used. To perform this selection, the ones with less than 100 instances were simply ignored, as done in the data set's original paper~\cite{zhu2016cvpr}. In addition, three other classes were removed. Those three categories (namely ``po'', ``io'', ``wo'') refer to three groups of signs that are not traffic signs, thus irrelevant. On GTSD, there are 43 unbalanced traffic sign classes. Commonly, the classification step used on this data set uses only three super-classes~\cite{wu_traffic_2013,garcia2018neurocomputing}: prohibitory, indicative and warning. In this work, the classification is performed using all 43 classes. In all datasets, only images containing traffic signs were used.

\subsection{Experiments}
Five experiments were carried out to evaluate the effectiveness of the proposed approach. The first was (i) an ablation study, followed by  (ii) training with a fully synthetic data set and (iii) search for model performance correspondence between training with real and synthetic images. Finally, two other applications were evaluated, using: (iv) the proposed method as data augmentation and (v) the proposed method for finetuning.

\subsubsection{Ablation Study}
\begin{figure}[h]
	\begin{center}
		\subfigure{\includegraphics[width=0.24\linewidth]{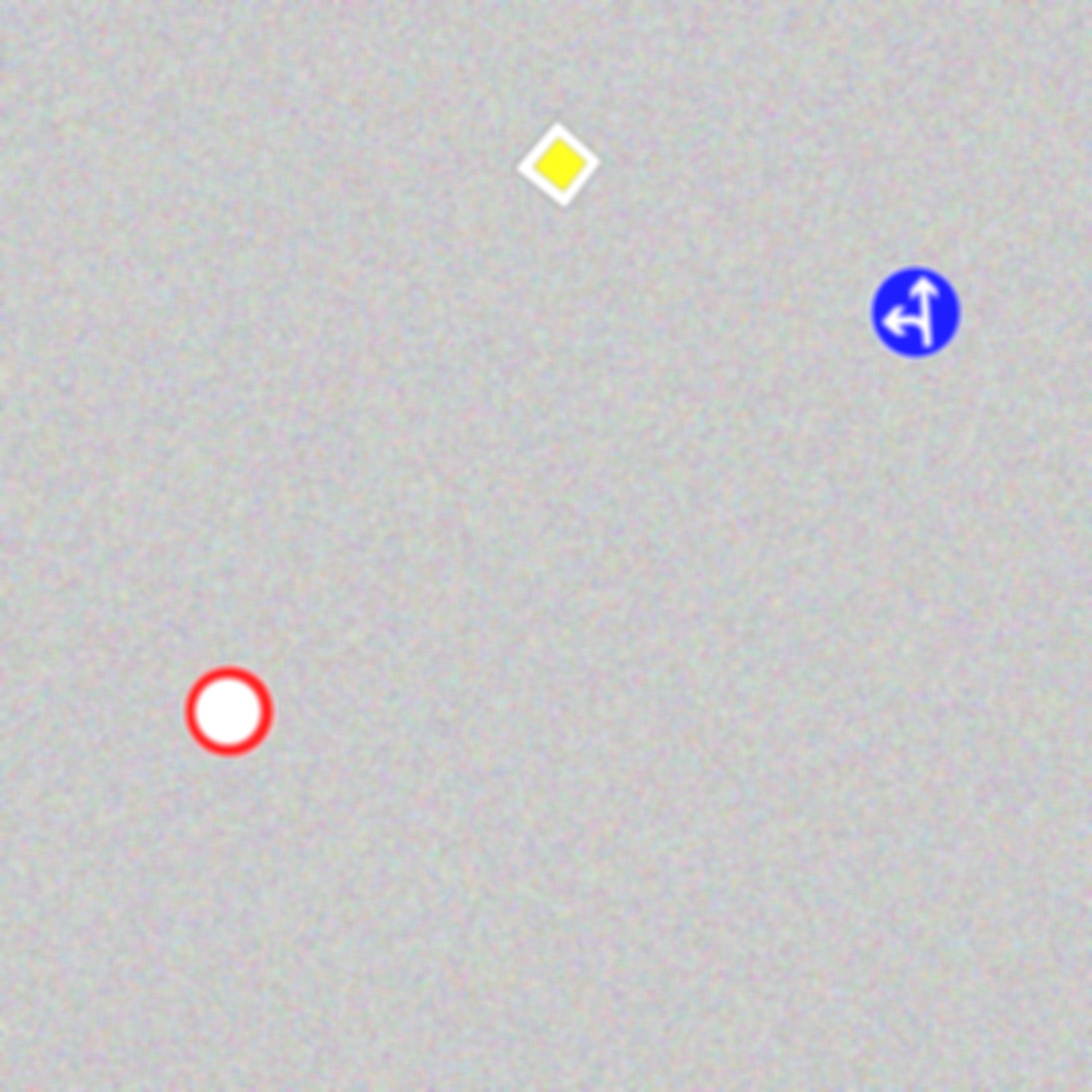}}
		\subfigure{\includegraphics[width=0.24\linewidth]{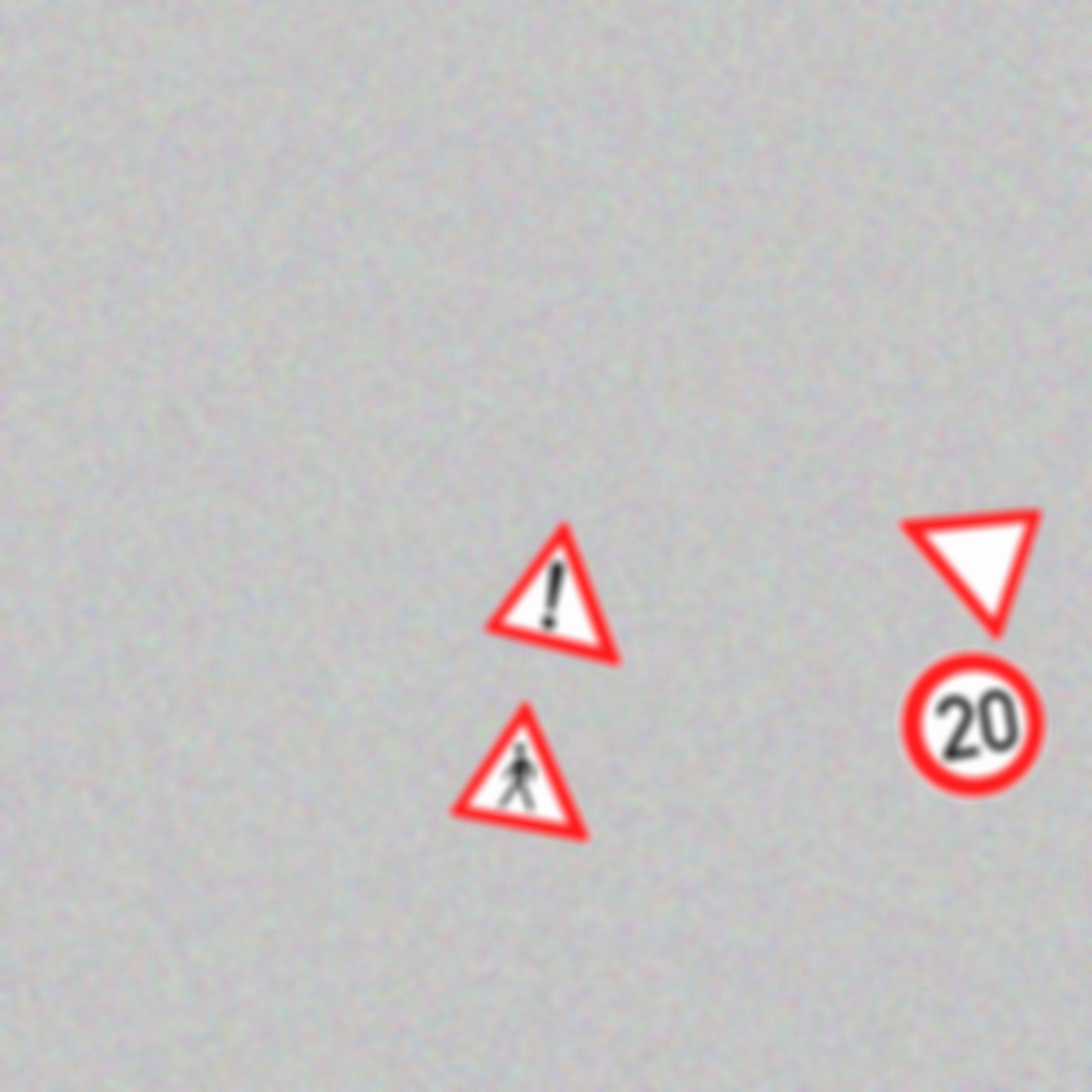}}
		\subfigure{\includegraphics[width=0.24\linewidth]{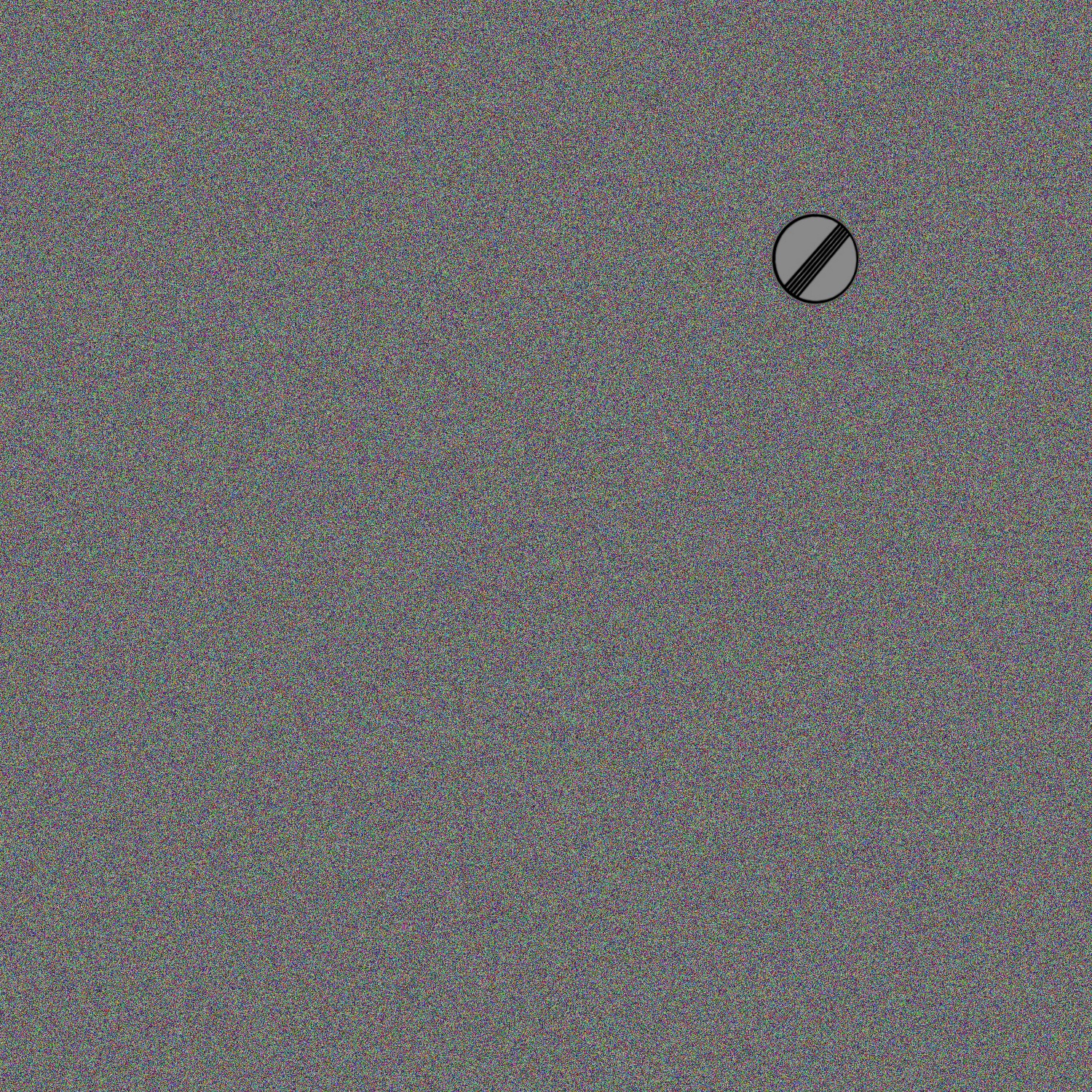}}
		\subfigure{\includegraphics[width=0.24\linewidth]{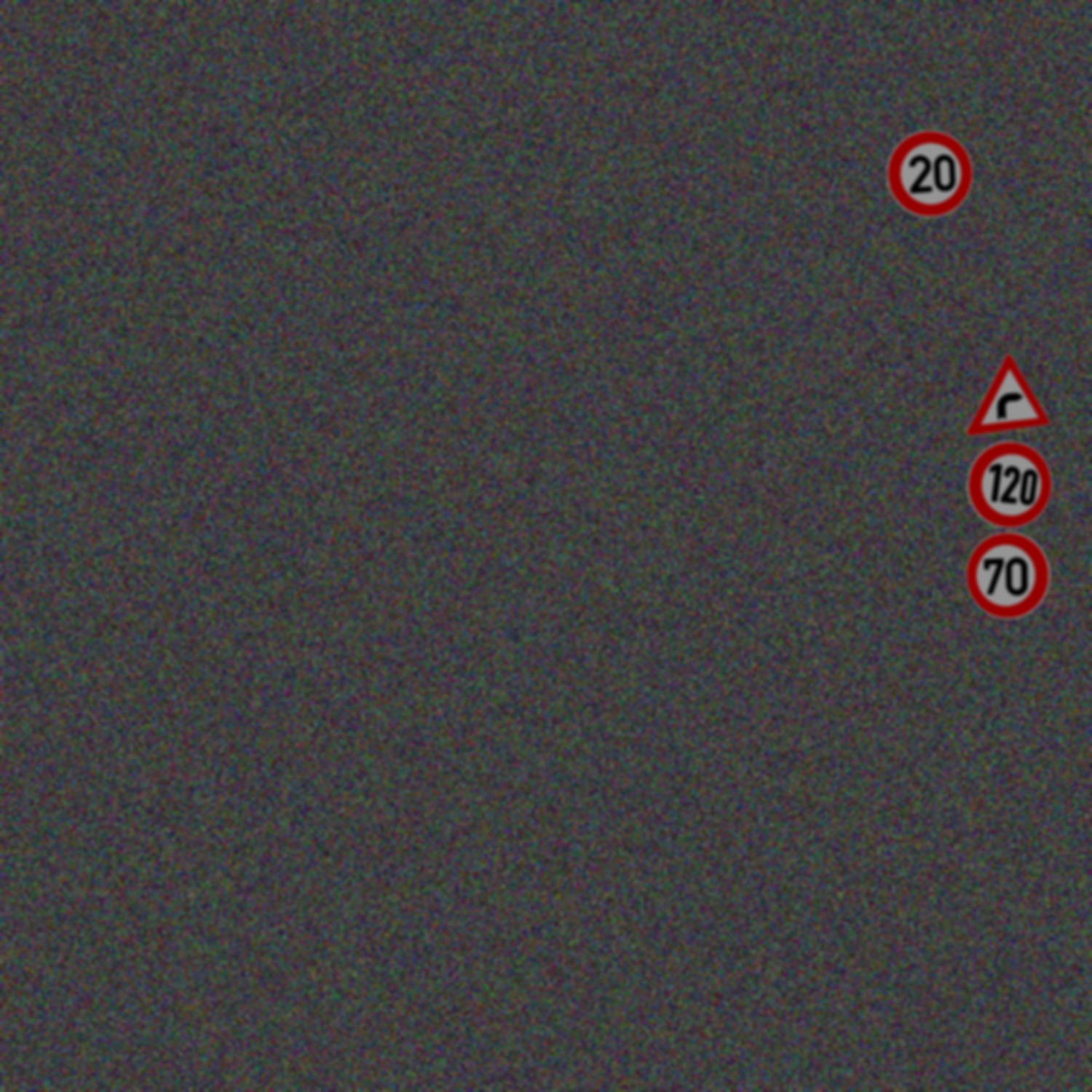}}
	\end{center}
	\caption{Four training samples with uniform noise as background for the ablation study.}
	\label{fig:noise_sample}
\end{figure}

The proposed approach to generate the training samples comprises a sequence of processes. To assess the importance of each process, the performance of the proposed approach was measured when disabling a process at a time. This study was conducted on the well-known GTSD data set. The study was performed on the following components from the training samples generation step: blur, brightness adjust, geometric transformations, background augmentation, blend, histogram noise and traffic sign grouping. Although other works have already shown that the Poisson blending algorithm~\cite{poisson} does not improve results in some problem domains~\cite{dwibedi_cut_2017}, it has not been shown yet on the traffic sign detection problem. Thus, the blending algorithm was tested as an alternative to the naive procedure adopted in our method. Two additional experiments were performed to verify the impact of using COCO's traffic scenes as background to train the deep detector: training with a data set generated using only images from the driving domain in COCO and with the full COCO data set as backgrounds. Additionally, training with a data set generated using images from the target data set that contains no traffic signs as background was also evaluated. This experiment was performed only on TT100K because GTSD does not provide images without traffic signs. The last experiment evaluates (for the three test collections) the replacing of COCO backgrounds by uniform random patterns. Some samples of training instances are shown in \Cref{fig:noise_sample}.

\subsubsection{Training with a Fully Synthetic Data set}

The main purpose of the proposed method is to be able to train a deep detector effortlessly, i.e., without manually annotating data, while maintaining its efficacy. For the evaluation, a deep detector was trained using only synthetic data generated by our method and compared against a model trained using only the full collection of real target-domain images (vanilla training). For fairer comparison, offline data augmentation was applied on the real image training set. This data augmentation comprises the same effects used on our method's background augmentation: brightness, contrast and blur. The total number of images in the training set after augmentation is the same as in training set with synthetic images (70k). This experiment aims to measure the performance cost of training only with samples generated by the proposed method instead of training only with real data of the target data set. To compare with the literature, the performance of two state-of-the-art methods, with objectives similar to ours, were also evaluated. In the first method~\cite{dwibedi_cut_2017}, objects of interest instances from the target data set's training set are cut and then pasted on images also from the target data set that contains no objects of interest. In the second~\cite{ferrari_modeling_2018}, a similar process is performed, but uses a neural network to predict regions that are likely to contain instances of each class. Then, it pastes instances from other images on those regions. Both  methods rely on segmentation annotations, which are much more expensive to acquire than bounding box annotations. As a consequence, they are evaluated only on TT100K since this data set is the only that provides segmentation annotations. Also, in both methods, the total number of training samples is 70k (same number as in every other training in this work).

\subsubsection{Model Performance Correspondence Between Training with Real and Synthetic Images}
\label{sec:exp_binary_search}
Another interesting question to answer is: How many real  images, manually annotated, are necessary to train a model that matches the performance of one trained with the proposed method? As an attempt to answer this question, an experiment was conducted using an approximate binary search in two data sets: BTSD and TT100K. GTSD was not evaluated because the performance with only synthetic images is already better than the vanilla training, as further discussed in \Cref{sec:results}. Due to time constraints, the number of splits in the binary search was set to a maximum of four for each data set, each split requiring a new training section. The search is approximate since: (i) there is randomness in the training step, (ii) the number of training sessions was limited to four (due to time restrictions), and (iii) the number of images used at each step does not match exactly the number given by the binary search. The latter is because of the method used to define the set of real image sets at each search step. This set is defined as $\mathcal{S} = \{S_{i}\}_{i=1}^{N}$, which is a set of $N$ sets of real images. Each set of images $S_{i}$ contains the same set of images as $S_{i-1}$ ($S_{0} = \emptyset$) plus $C$ more images selected in a way to maintain the original distribution of each class, where $C$ is the number of classes in the data set. This procedure was performed to reduce the randomness in the experiment. With the set of real images defined, data augmentation was applied to keep the number of training samples constant and equal to the number of samples used in the proposed method (i.e., 70k).

\subsubsection{Proposed Method as Data Augmentation}
One of the applications for the proposed method is data augmentation. To evaluate its effectiveness in this application, an experiment was performed on BTSD and TT100K. It was not performed on GTSD since it has not enough samples for each category. In this experiment, for each data set, Faster R-CNN was trained with a set of images  $\mathcal{I} = O\cup S\cup A$, where $O$ is a subset of the target country's training set, $S$ contains 35k synthetic images generated with the proposed method and $A$ contains $35\text{k} - \lvert O\rvert$ images generated using basic data augmentation (brightness, contrast and blur). In total, $\mathcal{I}$ contains 70k images. The data augmented is a set of images that contains at least $n$ original instances of each class. This approximate number is because it is very difficult to form a set with exactly $n$ instances per class, since an image can have multiple instances of different classes. Although it is not exactly $n$ instances, the set is chosen in a way that minimizes the difference. For each data set the experiment was performed with $n = 1, 10, M$ (maximum number possible, i.e., using the whole training set). To reduce the randomness, each set is a super-set of the previous, using the same procedure described on \Cref{sec:exp_binary_search}.

\subsubsection{Proposed Method for Finetuning}
The proposed method can also be used as a way to pretrain a model. After the pretraining, the model can be finetuned on a set of real images. In this experiment, for each data set, the  model trained with a fully synthetic data set (proposed method) was finetuned  on real images with simple augmentation (vanilla training). The inverse was also evaluated, i.e., training with real images with simple augmentation and then finetuning on the proposed method. In the finetuning process, the final learning rate of the pretrain ($10^{-4}$) was used during the whole training session, i.e., there was no learning rate decay.

\subsection{Faster R-CNN parameters}
Every training on Faster R-CNN used the same base settings, except for the input size for each data set. The feature-extractor used was the state-of-the-art ResNet-101~\cite{he2016deep}.
For tests on BTSD and GTSD, the proposed system was trained with an input size of 1500 $\times$ 1500 pixels, and tested with an input size of 3500 $\times$ 2060 pixels. The vanilla training for the GTSD data set was performed with an input size of 1360 $\times$ 800, while for BTSD the input size was 1628 $\times$ 1236. For TT100K, the input size was 2048$\times$ 2048 for training with both the proposed and vanilla methods. That is, all three vanilla trainings were executed in their respective original image resolutions. The input size for testing on TT100K was 4096 $\times$ 4096. For every target data set, the input size at test time for the proposed system and the vanilla trainings were the same. The increased test sizes are due to the poor performance of the tested detectors on small objects.

\subsection{Performance Metrics}

In addition to the precision and recall metrics, the Mean Average Precision (mAP) was also used to quantify the detection performance. The Average Precision (AP) metric, from which mAP is derived, follows the same approach in the PASCAL VOC 2012 challenge \cite{pascal-voc-2012}. Basically, AP is defined as the approximate area under
the precision/recall curve obtained for a fixed IoU threshold (0.5, in this work). Then, the mAP value is the
average of APs for all object classes.

\section{Results and Discussion}
In the following paragraphs, quantitative and qualitative results are presented and discussed following the same order of the respective experiments introduced in the last section. The section ends with a discussion of potential applications of the proposed method.

\label{sec:results}

\begin{figure*}[t]
	\centering
	\includegraphics[width=\linewidth]{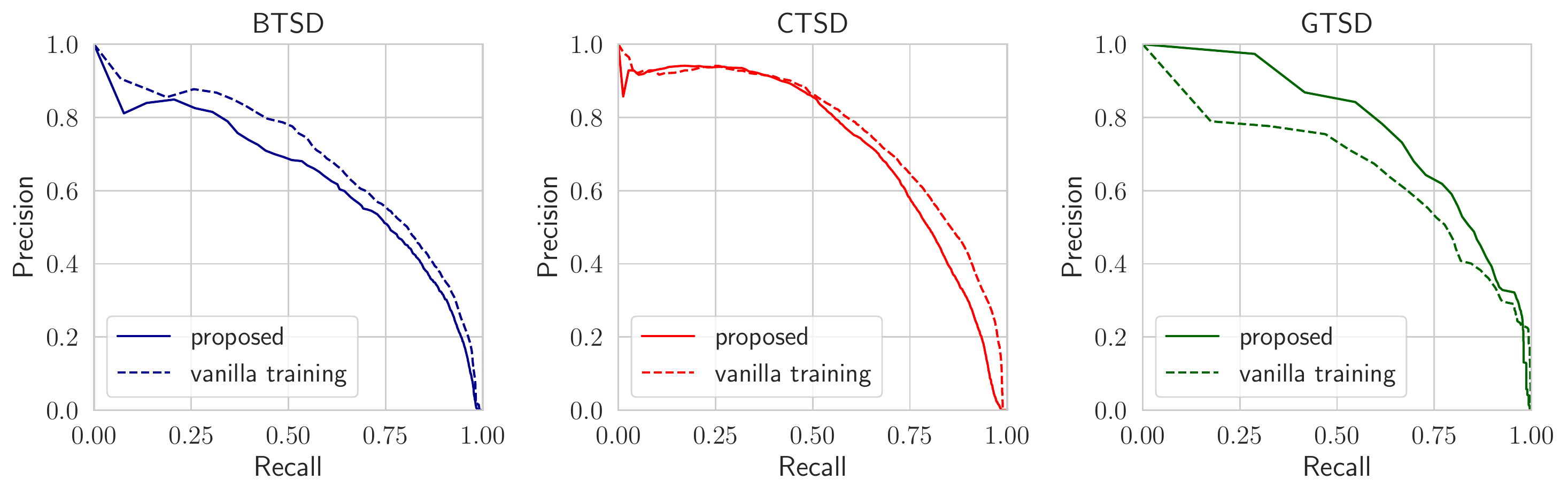}
	\caption{Precision-Recall curves for each data set.}
	\label{fig:precision_recall}
\end{figure*}

\begin{table}[]
	\begin{center}
		\caption{\label{tab:ablation_study}Ablation Study Results on GTSD}
		\begin{tabular}{lr}
			\toprule
			\textbf{Factor}              & \multicolumn{1}{l}{\textbf{mAP (\%) (normalized)}} \\ \midrule
			no brightness adjust         & 22.86 (0.2492)                                     \\
			no geometric transformations & 46.11 (0.5026)                                     \\
			no background augmentation   & 54.82 (0.5975)                                     \\
			no blur                      & 55.74 (0.6075)                                     \\
			no histogram noise           & 70.52 (0.7686)                                     \\
			no blend                     & 83.02 (0.9049)                                     \\
			Poisson blend                & 86.00 (0.9373)                                     \\
			no traffic sign grouping     & 86.77 (0.9457)                                     \\
			domain-only COCO backgrounds & 86.88 (0.9469)                                     \\
			no COCO filtering            & 87.01 (0.9483)                                     \\ \midrule
			none (full method)           & 91.75 (1.0000)                                     \\ \bottomrule
		\end{tabular}
	\end{center}
\end{table}
\vspace{.5em}
\PAR{Ablation Study.} The results for the ablation study are shown in \Cref{tab:ablation_study}. Overall, it can be observed that every component of the proposed method has a significant impact on the model's performance. Disabling the brightness adjustment was the step which yielded the worst performance, achieving only 24.92\% of the full method's performance, which may be a result of the high amount of brightness variations in the data set. Concurrently, traffic sign grouping had the lowest impact. Moreover, despite the simplicity of the blending process, results show that it substantially improves (9.51 p.p. more mAP) the performance of the detector on the target data set. The results with the Poisson blending indicate that a more complex blending process may not be necessary. In fact, it can worsen the model's performance, as the results suggest. Furthermore, the hypothesis that COCO driving domain images may hinder the learning process was also confirmed. In this scenario, the model's performance decays by 5.17\%.

The results for the study using uniform noise as background instead of real images are shown in \Cref{tab:noisebg}. In summary, they evidence that training with real images instead of noise indeed produces a better model, as expected. We observed, looking at the precision-recall curve, that the difference in mAP is mostly because of the increasing number of false positives. This behavior might be a result of the network being unable to learn features other than those in traffic signs, thus not learning well what a background is. On the other hand, if images that are known to not contain any traffic signs of interest are used, the performance increases by 2.12 p.p. on TT100K, but, in this case, more data were used.

\begin{table}[]
	\begin{center}
	\caption{\label{tab:noisebg} Results of training with uniform noise as background instead of COCO images on BTSD, TT100K and GTSD. $\Delta$ mAP corresponds to the difference when compared to training with COCO images as backgrounds.}
	\begin{tabular}{lll}
		\toprule
		\textbf{Data set} & \textbf{mAP (\%)} & \textbf{$\Delta$ mAP (\%)}      \\
		\midrule
		BTSD     & 71.61 & -8.67  \\
		GTSD    & 76.66 & -14.17 \\
		TT100K     & 63.99 & -19.13 \\
		\bottomrule
	\end{tabular}
\end{center}
\end{table}

\vspace{.5em}
\PAR{Training with a Fully Synthetic Data set.} Results of training with a fully synthetic data set generated by the proposed method are shown in \Cref{tab:comp-upperbound}, while precision vs. recall curves (as computed in~\cite{pascal-voc-2012}) are shown in \Cref{fig:precision_recall}. The results described as ``full method'' in \Cref{tab:ablation_study} are the same ones described as ``Proposed'' in \Cref{tab:comp-upperbound}. As evidenced by the GTSD results in the table, training with a fully synthetic data set may be even better than training with a smaller set of real images. For GTSD, the mAP difference between the proposed method and the vanilla training is +12.25 p.p.. Additionally, for BTSD and TT100K, which are larger data sets, the proposed method still performs similarly, although it does not surpass their corresponding vanilla trainings. The difference in mAP between the proposed method and the BTSD and TT100K vanilla trainings are -5.22 p.p. and -6.16 p.p., respectively. In \Cref{tab:comp-upperbound} the results of two state-of-the-art methods are also shown. Since both require real data to work, a fair comparison is against our result that uses real data, i.e., from the Proposed Method as Data Augmentation experiment. Our method outperforms both. This result shows that, for some problem domains, a more complex approach may not be necessary, in fact, it may produce a worse result.

As expected, increasing the collection of annotated traffic scenes yields a better performance, however this implies more human effort. Furthermore, it should be considered that the mAP difference between the vanilla training baseline and the proposed method is affected by the fact that  the baseline’s training and test sets share a particular geographic context. This implies more similarity between the two sets with respect to the overall appearance and the traffic scene structure, which is hard to be reached in real driving applications. By using natural images, the structural dependency is disregarded completely.

For a qualitative analysis, extensive qualitative results can be seen on the videos \footnote{\url{youtube.com/playlist?list=PLm8amuguiXiKEEI1A3qrktpup1SbmC1a5}}
Some of the false positives (i.e., false alarms) are indeed signs, but either they are not traffic signs, or they do not belong to set of the classes of interest. Furthermore, there is a considerable amount of false positives related to objects from the driving domain, such as headlights or traffic lights. This may be a result of the lack of those objects in the training set, although it can be mitigated by using real traffic scenes as backgrounds (as shown in the ablation study). Most false positives with higher confidence score arise from mistakes in the classification phase, but have sufficient IoU. Some false negatives (i.e., undetected traffic signs) can also be seen in the videos. The mistakes are mainly caused by severe geometric distortion or occlusion.

\begin{table}[]
	\begin{center}
	\caption{Comparison with state-of-the-art methods on BTSD, TT100K and GTSD. The "Proposed + real" results are from the Proposed Method as Data Augmentation experiment, using all real images available.}
	\label{tab:comp-upperbound}
	\begin{tabular}{@{}llrrr@{}}
		\toprule
		&                           & \multicolumn{3}{c}{\textbf{mAP (\%)}}                                                                     \\
		\textbf{Method} & \textbf{\begin{tabular}[c]{@{}l@{}}Real target\\domain images\end{tabular}} & \multicolumn{1}{l}{\textbf{BTSD}} & \multicolumn{1}{l}{\textbf{TT100K}} & \multicolumn{1}{l}{\textbf{GTSD}} \\ \midrule
		Vanilla                              & Required      & 85.50            & 89.28            & 79.50           \\ \midrule
		CPL~\cite{dwibedi_cut_2017}              & Required      & ---              & 89.66            & ---             \\
		CPL~\cite{dwibedi_cut_2017} + real       & Required      & ---              & 92.03            & ---             \\ 
		Context-DA~\cite{ferrari_modeling_2018}  & Required      & ---              & 74.96            & ---             \\ \midrule
		Proposed                                 & Not required  & 80.28            & 83.12            & \textbf{91.75}  \\
		Proposed + real                          & Required      & \textbf{89.64}   & \textbf{92.25}   & ---             \\ \bottomrule
	\end{tabular}
	\end{center}
\end{table}

\begin{figure}[h]
	\centering
	\includegraphics[width=\columnwidth]{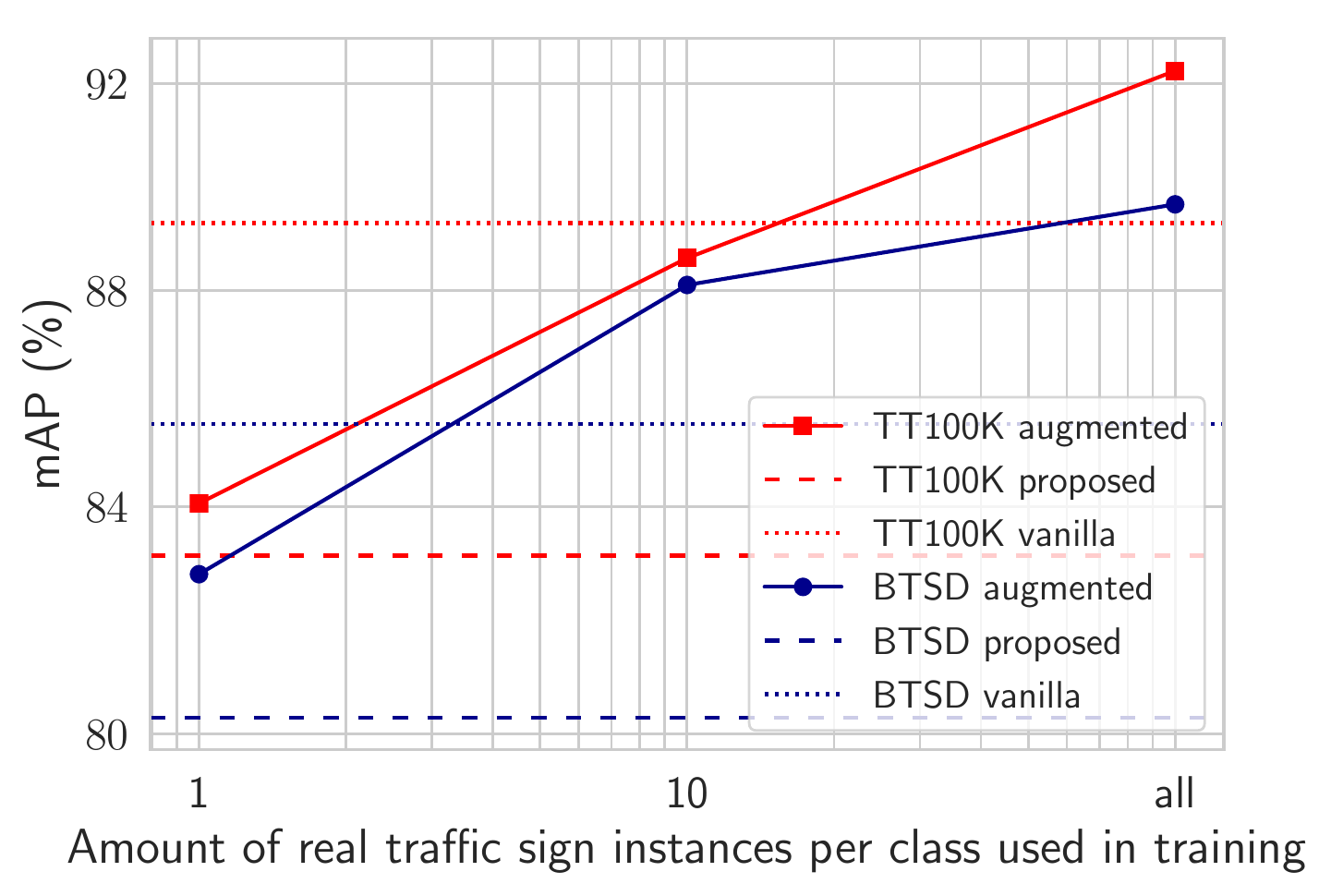}
	\caption{Proposed method as data augmentation (logarithmic scale).}
	\label{fig:aug}
\end{figure}

\vspace{.5em}
\PAR{Proposed Method as Data Augmentation.}
The results for the experiments using the proposed method as a data augmentation method are shown in \Cref{fig:aug}. For TT100K, using $n=10$ instances (at least) per class (total of 231 images), resulted in a mAP of only 0.67 p.p inferior to that obtained with the full training set that contains 5905 real images. For BTSD, training with also $n=10$ (total of 581 images), was enough to surpass the performance of training with the full training set that contains 4483 real images by 2.60 p.p.. Furthermore, when using the entire training set of real images ($n=M$), the proposed method was able to increase the model's performance significantly (2.97 p.p. and 4.41 p.p. of mAP, for TT100K and BTSD, respectively).

\vspace{.5em}
	\PAR{Finetuning on Real and on Synthetic Images.} The results for finetuning on real and on synthetic images are shown in \Cref{tab:finetuning}. As evidenced, using the proposed method to pretrain a model consistently increases its performance. When compared to the results with only an ImageNet pretrain, the mAP increases by 4.09 p.p., by 3.34 p.p. and by 18.98 p.p., for BTSD, TT100K and GTSD, respectively. Moreover, finetuning was slightly more effective than training with both data sets at the same time, as can be seen in \Cref{tab:comp-upperbound}, in the "Proposed + real" row. Those results show another possible use for the proposed method, as a cheap (without acquiring new real data) way to increase a model's performance. On the other hand, finetuning on synthetic data decreases the model's performance. This result is on par with the general intuition, as finetuning on synthetic images may specialize the model on them.

\begin{table}[]
	\begin{center}
	\caption{\label{tab:finetuning}Finetuning Results}
	\begin{tabular}{@{}lllr@{}}
		\toprule
		\textbf{Target} & \textbf{Training data set} & \textbf{Finetuning data set} & \multicolumn{1}{l}{\textbf{mAP (\%)}} \\ \midrule
		\multirow{2}{*}{BTSD}    & proposed method            & real + augmented           & \textbf{89.89}                                 \\
		& real + augmented        & proposed method               & 84.55                                 \\
		\multirow{2}{*}{GTSD}    & proposed method            & real + augmented           & \textbf{98.48}                                 \\
		& real + augmented        & proposed method               & 91.69                                 \\
		\multirow{2}{*}{TT100K}    & proposed method            & real + augmented     & \textbf{92.62}                                 \\
		& real + augmented        & proposed method               & 90.68                                 \\ \bottomrule
	\end{tabular}
	\end{center}
\end{table}

\begin{figure}[h]
\centering
\includegraphics[width=\columnwidth]{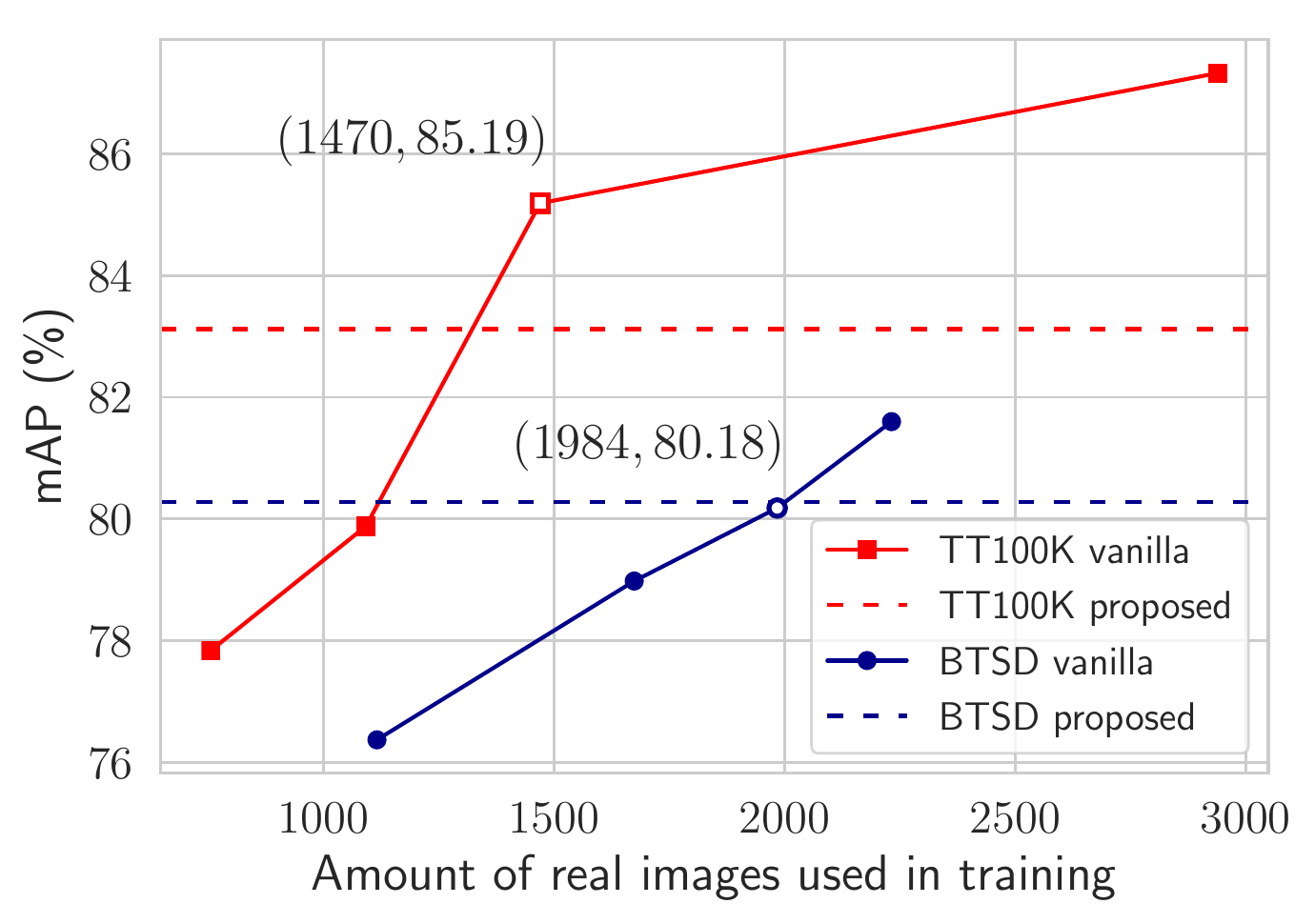}
\caption{Binary search for the equivalent amount of images on BTSD and TT100K. The highlighted points are the closest to the proposed method's result.}
\label{fig:binary-search}
\end{figure}

\vspace{.5em}
\PAR{Model Performance Correspondence Between Training with Real and Synthetic Images.}
\Cref{fig:binary-search} shows the results of the necessary number of real images to yield a performance comparable to that achieved by synthetically generated data. It is important to recall that the number of training samples was the same in each step of the search (70k). The difference is in the number of original real images (i.e., the rest of the images are generated via simple offline data augmentation). The results reveal that, for BTSD, the equivalence number is around 2000 images. For TT100K, this number is around 1250 images. For a simple comparison, to collect and annotate TT100K, 100k images had to be manually annotated, in order to acquire 10k images with traffic signs~\cite{zhu2016cvpr}. If the ten-to-one relation is kept, our method would eliminate the need to annotate around 12.5k images on TT100K. On BTSD, it would be equivalent to annotating around 20k images manually. This suggests that the proposed method may be specially useful when a large amount of annotated data is difficult to acquire.

\section{Conclusion}
Solving challenging tasks with deep neural network usually requires large-scale annotated data with real image samples belonging to the domain of the problem. The human effort and other costs involved in gathering such data has motivated research on alternative ways to train those models. In particular, this work leverages templates to train a deep model to detect traffic signs in real traffic scenes. Besides eliminating the need for real traffic signs, we also propose a more flexible and effortless construction of the training set by superposing the templates on natural images, i.e., arbitrary background images available in computer vision benchmarks. 
The results showed that the proposed method can be used to train deep detectors without the need for manually annotated data sets, and these models can achieve competitive performance. Moreover, multiple applications for the proposed method were shown, such as improving a model's performance when few, or even a lot of, annotated data is available. Finally, the number of real images that need to be manually annotated to match the performance of training with the proposed method was shown to be in the order of tens of thousands for every data set used. All those results indicate that the proposed method may make it viable to train a deep detector when large amounts of annotated data is difficult or impossible to acquire.

\ifCLASSOPTIONcompsoc
  \section*{Acknowledgments}
\else
  \section*{Acknowledgment}
\fi

The authors thank the NVIDIA Corporation for their kind donation of the GPUs used in this research. This study was financed in part by CAPES, FAPES, and CNPq.

\ifCLASSOPTIONcaptionsoff
  \newpage
\fi

\bibliographystyle{IEEEtran}
\bibliography{refs}

\begin{IEEEbiography}[{\includegraphics[width=1in,height=1.25in,clip,keepaspectratio]{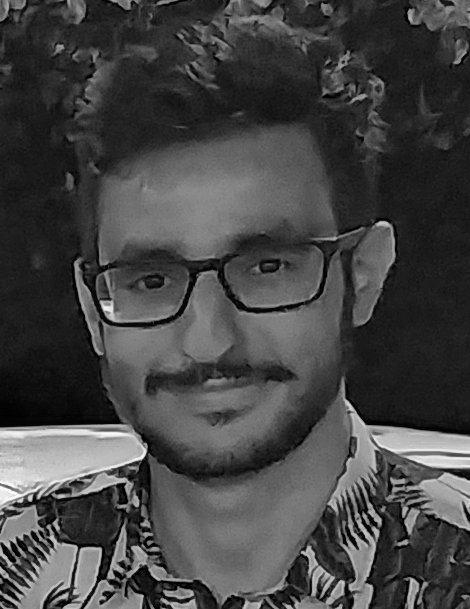}}]{Lucas Tabelini}
	is currently pursuing a B.S. degree in Computer Engineering at Universidade Federal do Espírito Santo (UFES), Brazil. Current research areas of interest include deep learning and computer vision.
\end{IEEEbiography}

\begin{IEEEbiography}[{\includegraphics[width=1in,height=1.25in,clip,keepaspectratio]{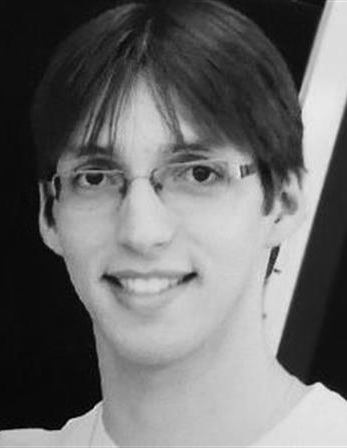}}]{Rodrigo Berriel}
received the M.Sc. degree in computer science in 2016 from the Universidade Federal do Espírito Santo (UFES, Brazil) and is currently pursuing his Ph.D. degree in the same university. His current research interests include computer vision and deep learning, particularly few-shot and lifelong learning.
\end{IEEEbiography}

\begin{IEEEbiography}[{\includegraphics[width=1in,height=1.25in,clip,keepaspectratio]{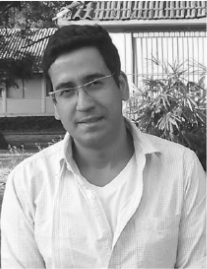}}]{Thiago M. Paixão}
	received the B.S. degree in computer science in 2007 from Universidade Federal de Minas Gerais (UFMG, Brazil), M.S. degree in computer science in 2010 from Universidade de São Paulo (USP, Brazil). He is currently pursuing the Ph.D. degree in computer science from Universidade Federal do Espírito Santo (UFES, Brazil). He is Professor at the Instituto Federal do Espírito Santo (IFES). His research interests include deep learning, computer vision, and structural pattern recognition.
\end{IEEEbiography}

\begin{IEEEbiography}[{\includegraphics[width=1in,height=1.25in,clip,keepaspectratio]{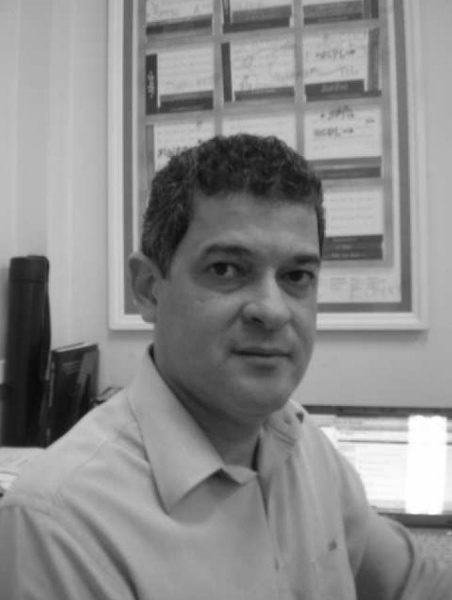}}]{Alberto F. De Souza}
	is a Full Professor of Computer Science and Coordinator of the Laboratório de Computação de Alto Desempenho (LCAD High Performance Computing Laboratory) at the Universidade Federal do Espírito Santo (UFES), Brazil. He received B. Eng. (Cum Laude) in electronics engineering and M. Sc. in systems engineering and computer science from Universidade Federal do Rio de Janeiro (COPPE/UFRJ), Brazil, in 1988 and 1993, respectively; and Doctor of Philosophy (Ph.D.) in computer science from the University College London, United Kingdom in 1999. Alberto F. De Souza is Senior Member of the IEEE and Comendador of the Rubem Braga Order.
\end{IEEEbiography}

\begin{IEEEbiography}[{\includegraphics[width=1in,height=1.25in,clip,keepaspectratio]{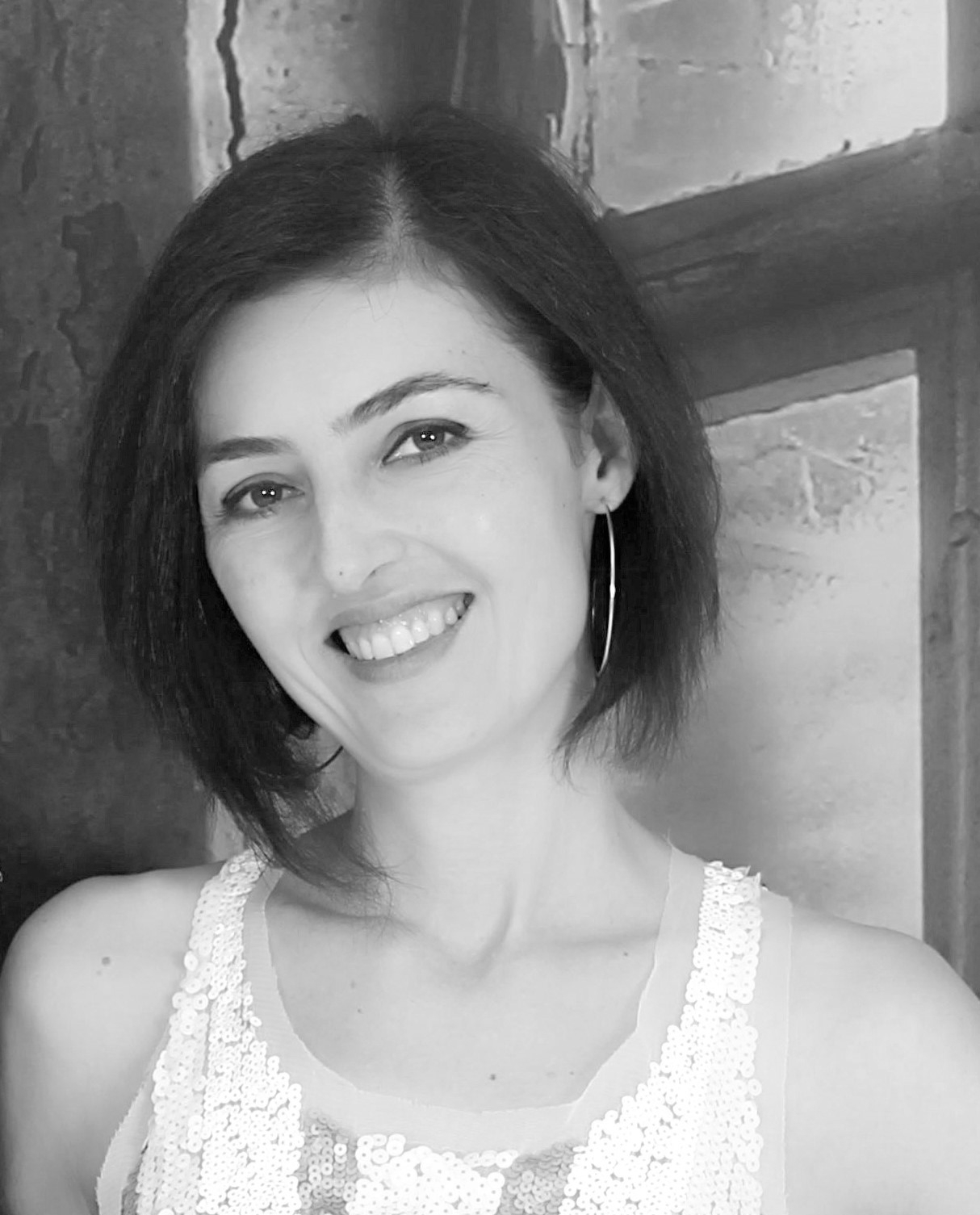}}]{Claudine Badue}
	is an Associate Professor of Computer Science and Co-Coordinator of the Laboratório de Computação de Alto Desempenho (LCAD High Performance Computing Laboratory) at the Universidade Federal do Espírito Santo (UFES), Brazil. She received a Bachelor’s degree in Computer Science in 1998 from the Universidade Federal de Goiás (UFG), a Ms.C. in Computer Science in 2001 and a Ph.D. in Computer Science in 2007, both from the Universidade Federal de Minas Gerais (UFMG). Her current research interests include autonomous cars, neural networks, and cognitive science.
\end{IEEEbiography}

\begin{IEEEbiography}[{\includegraphics[width=1in,height=1.25in,clip,keepaspectratio]{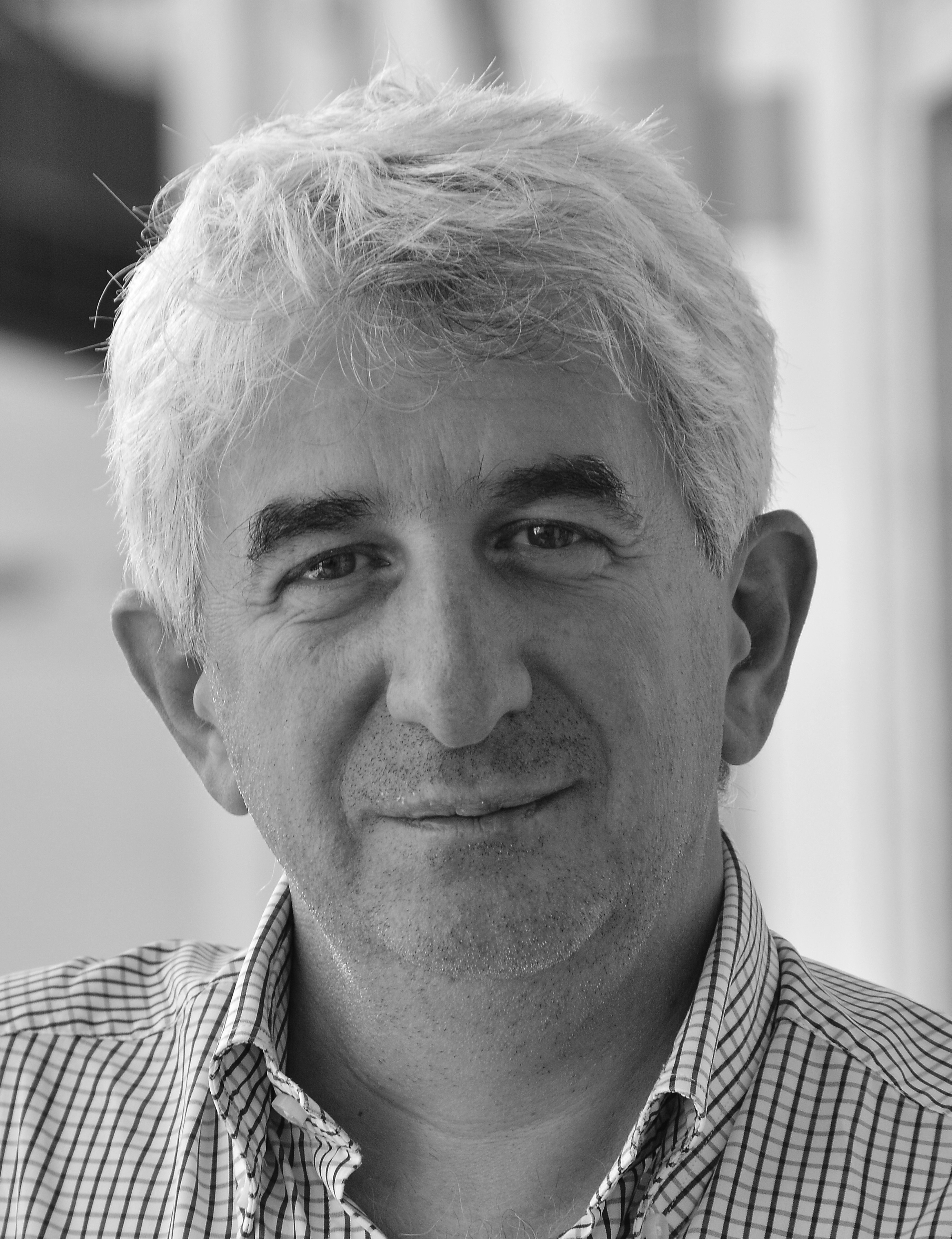}}]{Nicu Sebe}
	is Professor with the University of Trento, Italy, leading the research in  the areas of multimedia information retrieval and human behavior understanding.  He was the General Co-Chair of the IEEE FG Conference 2008 and ACM Multimedia 2013, and the Program Chair of the International Conference on Image and Video Retrieval in 2007 and 2010, ACM Multimedia 2007 and 2011. He was the Program Chair of ECCV 2016 and ICCV 2017, and a General Chair of ACM ICMR 2017. He is a program chair of ICPR 2020 and a general chair of ACM Multimedia 2022. He is a fellow of the International Association for Pattern Recognition.
\end{IEEEbiography}

\begin{IEEEbiography}[{\includegraphics[width=1in,height=1.25in,clip,keepaspectratio]{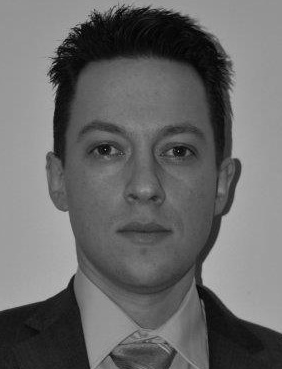}}]{Thiago Oliveira-Santos}
	received the B.S. degree in Computer Engineering in 2004 and the M.Sc. degree in Informatics in 2006 from Universidade Federal do Espírito Santo (UFES, Brazil). He received the Ph.D. degree in Biomedical Engineering in 2011 from Universität Bern (UNIBE, Switzerland). He is currently Professor at the UFES. Areas of interest include computer vision, image processing, computer graphics and image-guided surgery, cognition science, robotics.
\end{IEEEbiography}

\end{document}